\documentclass[12pt]{article}
\usepackage{graphicx}
\usepackage{amsmath, amsfonts, amssymb}
\usepackage{rotating}
\usepackage{multirow}

\title{Understanding and Reducing Crater Counting Errors in Citizen Science Data and the Need for Standardisation}
\author{P.D. Tar, N.A. Thacker}

\begin{document}

\maketitle

\begin{abstract}

Citizen science has become a popular tool for preliminary data processing tasks, such as identifying and counting Lunar impact craters in modern high-resolution imagery. However, use of such data requires that citizen science products are understandable and reliable. Contamination and missing data can reduce the usefulness of datasets so it is important that such effects are quantified. This paper presents a method, based upon a newly developed quantitative pattern recognition system (Linear Poisson Models) for estimating levels of contamination within MoonZoo citizen science crater data. Evidence will show that it is possible to remove the effects of contamination, with reference to some agreed upon ground truth, resulting in estimated crater counts which are highly repeatable. However, it will also be shown that correcting for missing data is currently more difficult to achieve. The techniques are tested on MoonZoo citizen science crater annotations from the Apollo 17 site and also undergraduate and expert results from the same region.

\end{abstract}

\section{Introduction}

Size-Frequency Distributions (SFDs) of impact craters are commonly used for investigating the evolution of planetary surfaces. A conventional SFD plots the cumulative frequencies of craters falling into geometrically increasing size-bands, normalised to a unit of surface area, thereby describing geological units in terms of crater diameters and densities \cite{Neukum1975}. Systematic crater surveys have been used to investigate Lunar history since the 1950s and 60s \cite{Hartmann1964}, with crater diameters in the order of a few km upwards being extensively catalogued. By the late 1970s the science of ``crater counting'' became an established method, incorporating the use of regression to fit functions describing crater production rates and surface chronology, with absolute surface ages being calibrated using Lunar samples \cite{Neukum1975}. Differences between earlier authors and proposed methods led to an advisory document \cite{CraterWorkingGroup} which made recommendations regarding the presentation and analysis of SFDs. Significant recommendations included the importance of understanding errors and also avoiding subjectivity where possible. In the spirit of these early recommendations this paper reassesses key error assumptions in an attempt to provide a framework for the \textit{objective} analysis of modern crater data gathered from high-resolution imagery.

\begin{figure} 
	\centering 
		\includegraphics[width=1.0\textwidth]{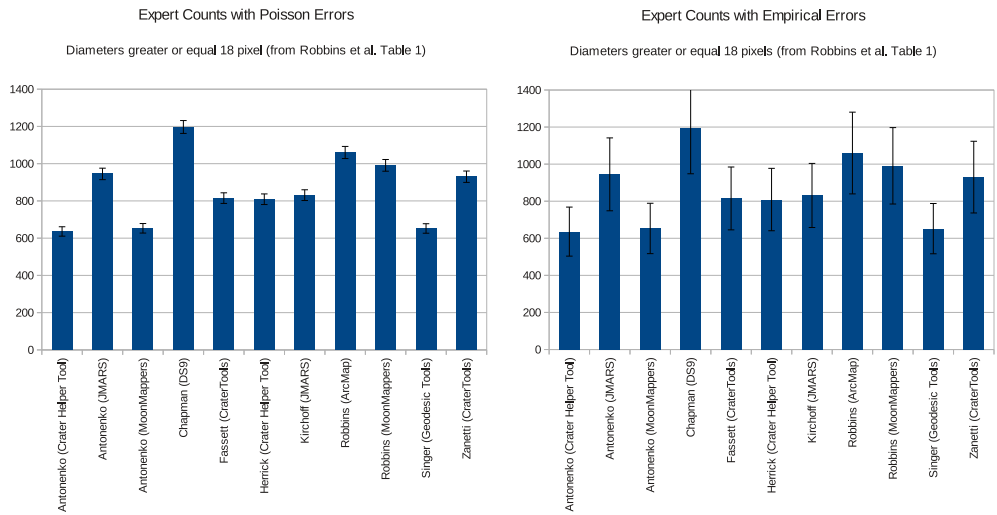}
	\caption{Expert counts from a common region should be statistically equivelent, i.e. equal to within the measurement accuracy. Repeatability data shows that Poisson errors are too small to account for differences between experts, even within a common counting region. Left: Sample of expert counts from Table 1, Robbins et al. 2014, with conventional Poisson $\sqrt{N}$ error bars; Right: Identical expert counts but with empirical error bars of 20.8\%, as reported by Robbins et al. showing true repeatability errors}
	\label{fig:non_poisson_errors}
\end{figure}

The availability of high-resolution Lunar images from Lunar Reconnaissance Orbiter Camera (LROC) \cite{Tooley} permits the study of small craters in the order of a few meters, rather than the kms of past decades. This data provides the potential for new insights into Lunar surface evolution at a finer scale than has previously been possible. However, manually identifying and counting the vast number of small craters present in modern images places a significant burden on the time and resources available to individual researchers. To mitigate against this, various automated solutions (e.g. \cite{Kamarudin} \cite{Kim} \cite{Bandeira}) and ``Citizen Science'' projects (e.g. \cite{Joy} \cite{Robbins}) have been proposed and tested. The quantitative use of crater statistics derived from these approaches relies heavily on developing a good understanding of the uncertainties in estimated counts. Conventional SFDs assume Poisson errors (i.e. $\sqrt{N}$) on crater counts and this entrenched assumption has become part of standard SFD analysis software \cite{Michael}. However, repeatability studies of experts and community crater counters \cite{Robbins2014} reveal uncertainties in counts between 5 to 7 times larger than those arising from Poisson perturbations alone (see figure \ref{fig:non_poisson_errors}). Empirical error rates used to assess automated crater detectors also reveal systematic effects which violate the simple Poisson assumption, i.e. false positive and negative detections. Despite this evidence, traditional Poisson error bars are still commonly used.

Perturbations in the ``true'' number of craters within a region can be described using Poisson statistics, which is justifiable due to the physical process of crater generation, i.e. rare impact events occurring in continuous time leading to variability across equivalent independent regions. This Poisson variability places a limit on the best accuracy achievable when counting craters, as equivalent independent regions will always be subject to these natural sampling errors. However, when making measurements within any specific region there are other sources of errors: false negatives, where genuine craters are not counted due to ambiguity or inattention; and false positives, where misidentified features that are not craters are mistakenly counted. Crater counts can thus be considered in these terms:

\begin{equation}
\label{equ:sfd_count_model}
N_D = N_T P_T + N_F P_F
\end{equation}

where $N_D$ is an estimated count of craters, either within a particular diameter range or a cumulative count; $N_T$ is the unknown ``true'' number of genuine craters within the counting region; $P_T$ is the fraction of craters actually identified and counted, falling between the values of zero and one; $N_F$ is the number of potentially ambiguous ``false'' features which might be mistaken for genuine craters; and $P_F$ is the fraction of the ``false'' features which are erroneously counted. To make quantitative use of crater counts researchers require honest estimates of errors, $\sigma_{N_D}$, so that meaningful comparisons between different SFDs or fitted functions can be made. Goodness-of-fit measures, such as K-S and $\chi^2$ statistics, also require error estimates to correctly normalise hypothesis or conformity tests. The use of such goodness-of-fits was recommended for confirming results of regression in \cite{CraterWorkingGroup}, variants of which assume Poisson errors, risking over-interpretation in the presence of larger amounts of uncertainty.

The total true error, $\sigma_{N_D}$, is a function of all terms (equ. \ref{equ:sfd_count_model}) contributing to the $N_D$ count. The values $N_T$ and $N_F$ can both be considered Poisson random variables. This is a reasonable assumption using the argument that features, be they craters or otherwise, are created through physical events in continuous time. The efficiency values $P_T$ and $P_F$ are also random variables. $P_T$ and $P_F$ are drawn from unknown distributions driven by psychological and perceptual factors in humans and by systematic biases in automated approaches. This paper proposes that it is the additional variability caused by these efficiency terms which boost crater counting errors, above Poisson, to the levels observed in practice in both humans and machines. Within this paper, a contamination estimation process to address the problem of false positive craters in MoonZoo data will be presented, i.e. for estimating and removing $N_F P_F$. A calibration process, which attempts to correct for missing craters $N_T (1-P_T)$, will also be proposed.

\begin{figure} 
	\centering 
		\includegraphics[width=1.0\textwidth]{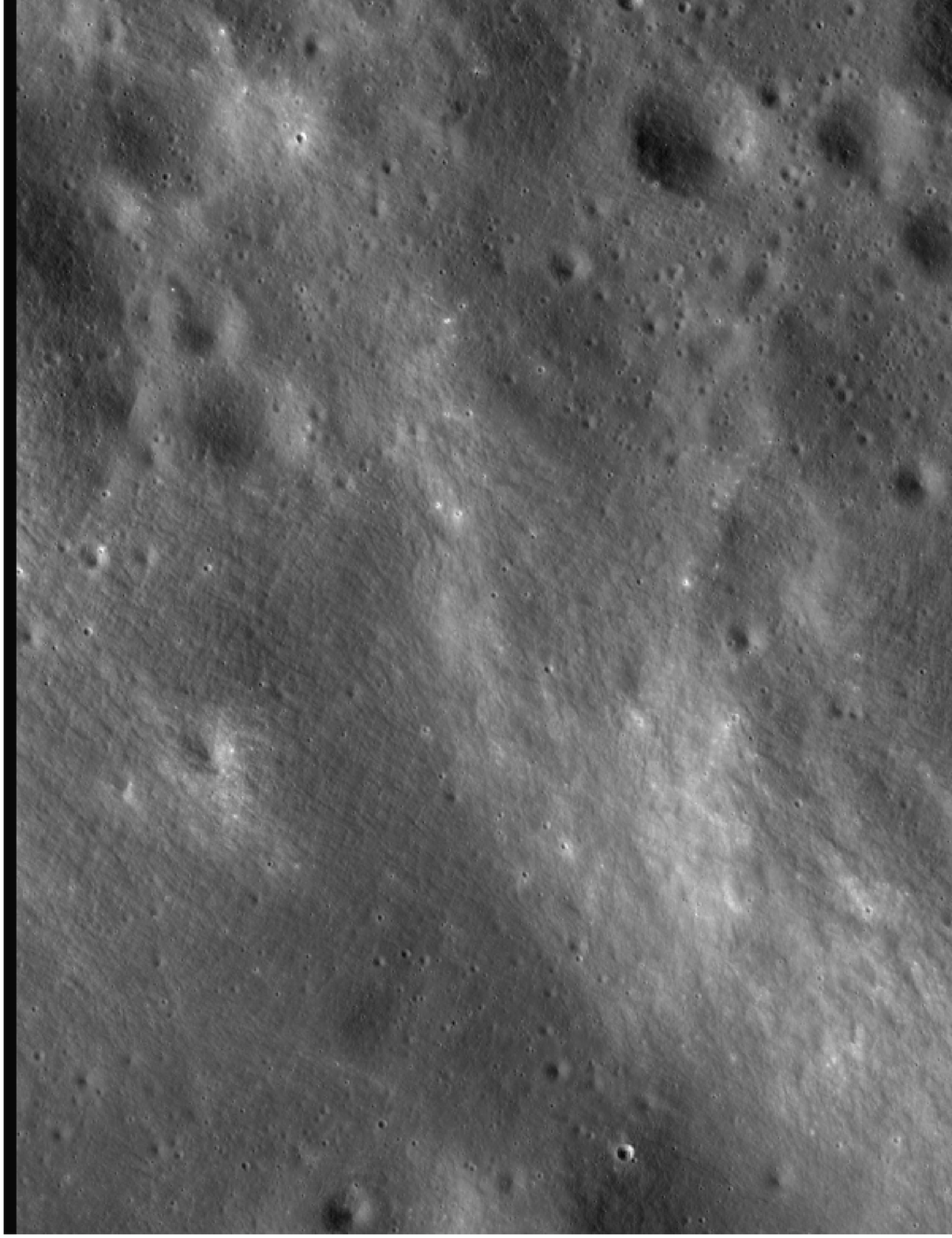} 
	\caption{Left: M104311715LE; Right: M104311715RE.} 
	\label{fig:mz_nacs}
\end{figure}

The Apollo 17 site has been extensively studied making it a useful target for advances in crater counting methods. In addition to physical samples and expert crater counts, the MoonZoo citizen science project has gathered 40,000+ crater annotations from this region generated by volunteers using a web-based interface\footnote{www.moonzoo.org}. The error analysis methods presented in this paper have been applied to this region, utilising crater counts from citizen scientists, with results checked against expert and undergraduate student counts. Reduced versions of the images used are shown in figure \ref{fig:mz_nacs}.

\section{Methodology}

The assessment of false positives and false negatives risks introducing circular arguments about the definitions of ``true'' and ``false'' craters, and also ``expert'' and ``non-expert'' counts. This risk is due to the requirement that some definition of what constitutes a crater is needed before the following processes can be applied. To prevent these circularities it is assumed that either a standard definition is available, or that one individual's subjective definition is applied consistently. The following methods can be understood as \textbf{objectively} applying an agreed upon \textit{subjective} definition, providing consistency and repeatability, which in the absence of any absolute ground-truth is arguably the best that can be achieved.

The false positive correction step, addressing $N_F P_F$, demonstrates the use of Linear Poisson Models (LPM) \cite{Tar} to make quantitative measurements. This is a supervised machine learning method which can learn classes of features (e.g. true and false craters) through appropriate examples, then estimate the amount of those features in new data. The false negative calibration step, addressing $N_T (1-P_T)$, demonstrates a simple scaling strategy for estimating the amount of missing data. These two steps are described separately below.

\subsection{False positive correction: addressing $N_F P_F$}
\label{sec:false_pos_correction}

The false positive correction process, which attempts to quantify the amount of contamination in crater counts, $N_F P_F$, requires the following information:

\begin{itemize}
\item a Lunar Reconnaissance Orbiter Narrow Angle Camera (NAC) image from which crater counts were derived;
\item x and y pixel coordinates of each candidate crater's centre in the coordinate system of the NAC image;
\item the diameter, in pixels, of each candidate crater;
\item and a ``standard'' crater definition in the form of a template image.
\end{itemize}

GIS systems can easily convert between pixel coordinates and geo-coordinates, so for convenience of implementation the proposed method makes use of pixel space information only. In the case of MoonZoo data, the x, y and diameter parameters are provided by a clustering algorithm which coalesces multiple citizen science mark-ups into individual candidate craters [cite roberto and epsc2014]. Other sources of candidate craters are also possible, including outputs from algorithms or ensembles of expert annotations. After identifying a representative subset of ``true'' and ``false'' craters which have the same illumination conditions as the target crater counting region, the false positive correction process involves:

\begin{enumerate}
\item computing a template crater image from the ``true'' subset;
\item using a carefully constructed similarity measure, comparing the template crater against all ``true'' and ``false'' examples and recording the resulting match scores;
\item populating histograms with the distribution of the match scores for ``true'' and ``false'' examples;
\item using the histograms to train a Linear Poisson Model (LPM), which results in a set of Probability Mass Functions (PMFs) which can be combined linearly to describe future data;
\item using Extended Maximum Likelihood to fit the PMFs to the candidate crater match scores requiring correction;
\item interpreting the LPM's linear weighting factors to determine ``true'' and ``false'' crater quantities;
\item using the LPM's error theory to provide predictive error bars on the remaining ``true'' crater quantities.
\end{enumerate}

A LPM can be used to model arbitrary low-dimensional data distributions, as long as they can be modelled as linear combinations of histograms. The histograms must encode image information in order to differentiate between true and false craters, yet there are numerous ways this information could be encoded. To show that the method can produce statistically equivalent outputs independently of the encoding, two different types of crater template and two different similarity scores will be used to populate histograms. Whilst this does not cover every possible encoding scheme, it is sufficient to demonstrate the generic nature of the approach.

\subsubsection{Crater template construction}
\label{sec:crater_template_construction}

\begin{figure} 
	\centering 
		\includegraphics[width=1.0\textwidth]{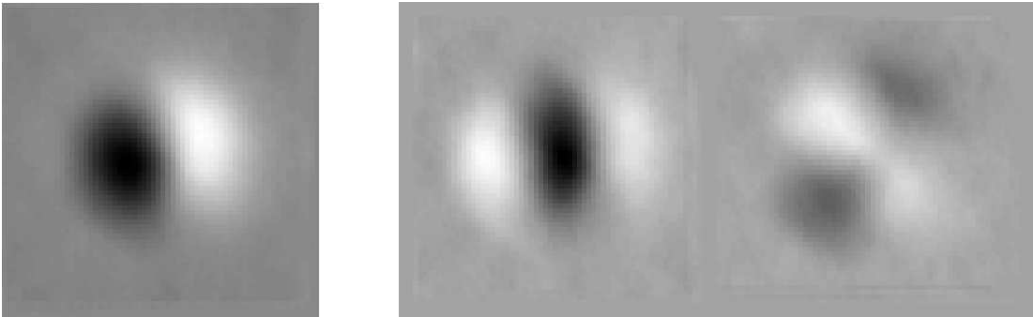} 
	\caption{Left: Mean grey level crater template derived from MoonZoo data. Right: Combined horizontal and vertical gradient (x, y derivative) template.} 
	\label{fig:mz_templates} 
\end{figure}

Two different forms of template have been investigated: an average crater appearance with mean local illumination removed; and a derivative template modelling changes in illumination both horizontally and vertically across an average crater. Examples of these templates can be seen in figure \ref{fig:mz_templates}. Both types of template begin with multiple crater image examples scaled to a common size. For each example, the mean grey level of the image is subtracted from each pixel to remove the effects of local illumination and albedo. The average template is given by the mean per-pixel values computed across all examples. The derivative template is given by the mean pixel differences between adjacent pixels to the left and right, and above and below, with the horizontal and vertical results concatenated next to one another. The derivative template further reduces the effects of local illumination and albedo variations by focusing on differences rather than absolute values.

\subsubsection{Similarity measures}
\label{sec:similarity_measures}

Two different forms of similarity measure have also been investigated: the mean of squared residuals between image and template pixels (i.e. mean squared error, $S_{MSE}$); and a normalised dot-product which treats image and template pixels as vectors ($S_{DP}$). Both of these measures are motivated by a Likelihood interpretation of template matching which assumes independent Gaussian noise on image pixels:

\begin{equation}
\mathcal{L}_{match} = \prod_i^n e^{-(a_i-b_i)^2}
\end{equation}

where $a$ is the template; $b$ is the image patch being matched; $i$ is an index over each pixel; and $n$ is the number of pixels in the template. It can be seen that maximising this Likelihood is equivalent to minimising $S_{MSE}$:

\begin{equation}
S_{MSE} = \frac{1}{n} \sum_i^n (a_i-b_i)^2
\end{equation}

\begin{equation}
\ln \mathcal{L}_{match} = -\sum_i^n (a_i-b_i)^2
\end{equation}

where the peak in the log Likelihood coincides with the peak in the Likelihood and $S_{MSE} \propto - \ln \mathcal{L}_{match}$. It can also be seen that maximising $S_{DP}$ is approximately equivalent to minimising $S_{MSE}$ by inspecting:

\begin{equation}
S_{DP} = \frac{1}{n \|a\|} \sum_i^n a_i b_i
\end{equation}

\begin{equation}
S_{MSE} = \frac{1}{n} \left( \sum_i^n a_i^2 + \sum_i^n b_i^2 - 2\sum_i^n a_i b_i \right)
\end{equation}

where $\sum_i^n a_i^2$ and $\|a\|$ are constant for a fixed template; and $\sum_i^n b_i^2$ is dominated by the mean grey level of the image patch being matched, which is approximately constant within a local image region. The $S_{PD}$ similarity measure therefore focuses on matching the high spatial frequency components of the crater templates which provides further invariance to local illumination and albedo effects.

\subsubsection{Applying the templates}
\label{sec:applying_the_templates}

The construction of templates and selection of similarity measures presented above assume the only sources of variability are local illumination conditions and pixel-level noise. However, the degradation state of craters also significantly affects appearance. To approximately accommodate the effects of degradation, images being matched are first smoothed using Gaussian blurring, as the blurring of a crater image visually mimics the effects of erosion, allowing for improved matches. The match score assigned to a given candidate crater is then computed as follows:

\begin{enumerate}

\item smooth the crater image by a small amount;
\item subtract the mean local grey level;
\item compute horizontal and vertical derivatives if the derivative template is to be used;
\item compare the template to the crater using one of the similarity measures;
\item repeat the process for different smoothing levels until the best match score is achieved;

\end{enumerate}

The best match scores achieved per crater are recorded and their distributions are accumulated into histograms. The histograms of match scores for verified ``true'' and ``false'' craters are used as definitions against which histograms of unknown crater match scores can be subsequently assessed.

\subsubsection{The application of LPMs}
\label{sec:application_of_LPMs}

A LPM \cite{Tar} can describe the shape and variability found within histograms using a linear combination of simpler fixed components:

\begin{equation}
\mathbf{H} = \mathbf{P} \mathbf{Q} + \mathbf{e}_H 
\end{equation}

where $\mathbf{H}$ is a histogram, with elements $\mathbf{H}_X$; $\mathbf{P}$ is an $m$ by $n$ matrix describing the Probability Mass Functions (PMFs) of $n$ components with elements $\mathbf{P}_{ij} = P(X=i|k=j)$, i.e. the probability of an entry in bin $X$ (i.e. a match score range) given component $k$; $\mathbf{Q}$ is a column vector of $n$ quantities corresponding to the amount of each component present within the histogram; and $\mathbf{e}_H$ is a column vector of noise assumed to be independent Poisson perturbations consistent with typical histogram formation. The inverted formulation, which is appropriate for making quantity measurements, is then given by: 

\begin{equation} 
\label{equ:Q_invPH} 
\mathbf{Q} = \mathbf{P}^{-1} \mathbf{H} + \mathbf{e}_{Q} 
\end{equation} 

where $\mathbf{P}^{-1}$ is an $n$ by $m$ matrix with elements consistent with Bayes Theorem, $\mathbf{P}^{-1}_{ij} = P(k=i|X=j)$, i.e. the probability that component $k$ was the source of an entry in bin $X$; and $\mathbf{e}_{Q}$ is noise on the quantities.

During training, a LPM must determine the necessary PMFs required to describe the distribution of ``true'' and ``false'' crater match scores. Once these have been established they can be fitted to new histograms containing unknown quantities of contamination, thus estimating how much of each category exists within the data. Both training and fitting are achieved using Expectation Maximisation \cite{Dempster} to optimise the following Extended Maximum Likelihood:

\begin{equation}
\ln \mathcal{L} = \sum_X \ln \left[\sum_k P(X|k)\mathbf{Q}_k\right] \mathbf{H}_X - \sum_k \mathbf{Q}_k 
\end{equation}

During training, this function is jointly optimised for a set of example histograms giving a set of $P(X|k)$ components. This is performed separately for the ``true'' and ``false'' classes, resulting in a set of PMFs associated with each class. The number of components required to describe each class is determined by adding additional components until the $\chi^2$ per degree of freedom between LPM and example histograms approaches unity. During contamination estimation in new data, this function is optimised to fit true/false classes of component by adjusting weighting quantities, $\mathbf{Q}_k$, which are then summed within their respective classes to give total quantities of ``true'' craters and ``false'' contamination.

Sampling errors in training histograms and incoming data combine to give a level of uncertainty on the estimated quantities. In order to factor these uncertainties into final crater counts they must be propagated through the EM algorithm using error propagation \cite{Barlow}:

\begin{equation} 
\mathbf{C}_{Q} = \mathbf{C}_{data} + \mathbf{C}_{model} 
\end{equation} 

\begin{equation} 
\mathbf{C}_{ij(data)} = \sum_X \left [ \left (\frac{\partial \mathbf{Q}_i}{\partial \mathbf{H}_{X}} \right ) \left ( \frac{\partial \mathbf{Q}_j}{\partial \mathbf{H}_{X}} \right ) \sigma^2_{\mathbf{H}_{X}} \right] 
\end{equation} 

\begin{equation} 
\mathbf{C}_{ij(model)} = \sum_X \left[ \sum_k \left ( \frac{\partial \mathbf{Q}_i}{\partial \mathbf{H}_{X|k}}  \right ) \left ( \frac{\partial \mathbf{Q}_j}{\partial \mathbf{H}_{X|k}}  \right )  \sigma^2_{\mathbf{H}_{X|k}} \right ] 
\end{equation} 

where $\mathbf{C}_{Q}$ is the error covariance matrix for the estimated quantities; $\mathbf{C}_{data}$ is the statistical contribution of the error from the incoming histogram data; and $\mathbf{C}_{model}$ is the systematic contribution from the training exemplar histograms used to construct the LPM. These form the basis of SFD error bars.

\subsection{False negative calibration: addressing $N_T (1-P_T)$}
\label{sec:false_neg_calibration}

The false negative calibration process, which attempts to correct for uncounted craters, $N_T (1-P_T)$, requires the following information:

\begin{itemize}
\item a subset of diligently estimated ``expert'' crater counts covering regions which overlap the MoonZoo crater counting region;
\item an estimate of the ``expert'' repeatability;
\item and a MoonZoo crater count which has already been corrected for contamination from false positives.
\end{itemize}

The process of calibration involves:

\begin{enumerate}

\item estimating corrective scaling factors based upon the ratio of ``expert'' to non-expert counts within the overlapping calibration regions;
\item applying the scaling factors to all non-expert counts;
\item using error propagation to apply the scaling factors to the errors on non-expert counts;
\item assessing the success of the calibration using a $\chi^2$ per degree of freedom check on the calibrated regions.

\end{enumerate}

Scaling factors can be computed as an overall correction, a diameter specific correction, a degradation specific correction etc. as required. A scaling factor, $s$, can be computed to approximately correct for missing data:

\begin{equation}
\label{equ:scaling_factor}
s = \frac{u}{m_0}
\end{equation}

where $u$ is a ``ground truth'' frequency within an SFD bin, as set by some standard or expert; and $m_0$ is the predicted frequency of true positives from the previous correction stage. A false negative corrected count for future data is then be given by:

\begin{equation}
\label{equ:corrected_count}
c = ms
\end{equation}

where $m$ is a false positive corrected count from a different area. The error on this new count, $\sigma_c^2$, is given by:

\begin{equation}
\label{equ:corrected_var}
\sigma_c^2 = s^2 \sigma_m^2 + m^2 \sigma_s^2
\end{equation}

where $\sigma_m^2$ is the variance on $m$, as given by the LPM error theory; and $\sigma_s^2$ is the variance on the scaling factor, given by:

\begin{equation}
\label{equ:scaling_factor_var}
\sigma_s^2 = \frac{1}{m_0^2} \sigma_u^2 + \frac{u^2}{m_0^4} \sigma_{m_0}^2
\end{equation}

where $\sigma_u^2$ is the estimated error on the ``ground truth'' count, which must be determined from repeatability data, e.g. the repeatability of a single expert when counting the same area multiple times.

The effectiveness of the calibration can then be checked using a chi-square per-degree of freedom test, which should be unity if corrected SFDs match experts within allowable margins of error:

\begin{equation}
\label{equ:chi_test}
\chi^2_d = \frac{1}{d} \sum_i \frac{(c_i - u_i)^2}{\sigma^2_c + \sigma^2_u}
\end{equation}

where $c_i$ is a corrected count in range $i$; $u_i$ is an ``expert'' count; and $d$ are the number of degrees of freedom in the model.

\section{Experiments}

The two correction steps have been tested using craters annotated by MoonZoo users around the Apollo 17 site. The false positive correction step has been tested using different combinations of free parameters in order to show that statistically equivalent results can be achieved using LPMs with different input representations. The false negative calibration step has been tested using an overall scaling factor and also size specific scaling factors. These steps are described separately below.

\subsection{False positive correction}

Approximately 20,000 candidate craters from MoonZoo with diameters of 20 pixels or greater (clustered from 40,000+ raw annotations [cite MoonZoo 1 paper]) were individually assessed and categories into ``true'' and ``false'' examples. These were derived from the following NAC images:

\begin{itemize}
\item M104311715LE
\item M104311715RE
\end{itemize}

which are both approximately 5,000 by 50,000 pixels in size. These were used to train LPMs to test the false positive contamination correction. Templates were created (section \ref{sec:crater_template_construction}) from candidate craters which were annotated by 3 or more MoonZoo users (which all fell into the category of ``true'' craters). These templates were 60 by 60 pixels in size (120 by 60 for the concatenated derivative template) with all example craters scaled to a 40 pixel diameter centred within the templates. One dimensional match score distributions (section \ref{sec:similarity_measures}) were created for the four possible combinations of templates and similarity measures:

\

\begin{tabular}{|c|c|c|}
\hline 
• & Avg Appearance (Grey level) & Avg Derivative (Gradient) \\ 
\hline 
$S_{MSE}$ & Grey MSE & Grad MSE \\ 
\hline 
$S_{DP}$ & Grey DP & Grad DP \\ 
\hline 
\end{tabular} 

\

Two dimensional match score distributions were also created using combinations of the above, giving:

\

\begin{tabular}{|c|c|c|c|}
\hline 
• & Grey MSE & Grad MSE & Grey DP \\ 
\hline 
Grad MSE & X & • & • \\ 
\hline 
Grey DP & X & X & • \\ 
\hline 
Grad DP & X & X & X \\ 
\hline 
\end{tabular} 

\

These different histograms each contain subtly different information regarding the image evidence used to differentiate between true and false craters. By applying LPMs to each of these possible combinations it is possible to: \textbf{a)} demonstrate the consistency of the method by showing that equivalent results can be achieved for the different evidence used; \textbf{b)} select the most efficient combination to achieve best absolute levels of counting repeatability.

To achieve best template matches 16 logarithmic levels of image smoothing (section \ref{sec:applying_the_templates}) were employed. The Gaussian smoothing filter widths used, in pixels, were:

\

\begin{tabular}{|c|c|c|c|c|c|c|c|}
\hline 
0.10 & 0.12 & 0.14 & 0.17 & 0.21 & 0.25 & 0.30 & 0.36 \\ 
\hline 
0.43 & 0.52 & 0.62 & 0.74 & 0.89 & 1.07 & 1.28 & 1.54 \\ 
\hline 
\end{tabular} 

\

LPMs (section \ref{sec:application_of_LPMs}) were repeatedly constructed and applied to estimate the quantity of ``true'' and ``false'' craters in randomised samples, with 1,000 repeated measurements made in order to produce predicted and empirical error distributions on the estimated quantities. During each trial craters were selected in the form of random rectangular regions sampled with replacement. Regions rather than individual craters were selected to preserve any local spatial correlations and sampling with replacement was used to achieve the required quantity of data for thorough testing. A range of data quantities were tested, using 0.01, 0.10, 1.00, 10.00 and 100.00 times as much testing data as training data. The quantity of contamination within each trial matched that found within the raw MoonZoo data, which is approximately one quarter.

After each trial the difference between known ground-truth values and estimated values were divided by the predicted error and recorded. The predicted accuracies were also recorded as percentage errors on measured quantities.

\subsection{False negative calibration}

Crater counts were provided by undergraduate students, under supervision, down to craters with 20 pixel diameters for the following NAC image strips:

\begin{enumerate}
\item Image M104311715LE, pixel rows 400 to 800
\item Image M104311715LE, pixel rows 6,000 to 6,400
\item Image M104311715LE, pixel rows 15,200 to 15,600
\item Image M104311715LE, pixel rows 32,000 to 32,400
\item Image M104311715LE, pixel rows 49,500 to 49,900
\item Image M104311715RE, pixel rows 400 to 800
\item Image M104311715RE, pixel rows 6,000 to 6,400
\item Image M104311715RE, pixel rows 15,200 to 15,600
\end{enumerate}

which were used to test the false negative missing data correction. Each strip was annotated by two different students to provide repeatability data. The variance on the repeatability was computed by fitting a Binomial distribution to the repeatability data, fixing the number of Binomial trials to the total number of craters and computing the success probability to match the observed frequencies of single and double mark-ups, i.e. how many of the craters were only annotated by only one student against how many were annotated twice. Error bars on expert counts were then computed using a Binomial variance with the same success probability:

\begin{equation}
\sigma^2 = u (P - P^2)
\end{equation}

where $u$ is an undergraduate count and $P$ is the success probability.

Expert crater counters were also provided (need details).

Overall scaling factors were computed and also diameter dependent scaling factors. As a self-consistency check, the scaling factors were applied back in the calibration regions (i.e. a best case, most representative scenario) and $\chi^2$ per degree of freedoms computed to assess the quality of fit between corrected MoonZoo counts, undergraduate counts and expert counts.

\section{Results}

Results for the two different processing steps are presented separately below.

\subsection{False positive correction}

\begin{figure} 
	\centering 
		\includegraphics[width=\textwidth]{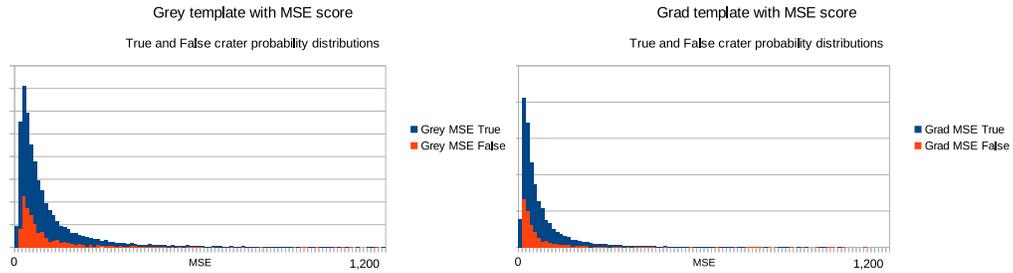} 
	\caption{Left: Mean Squared Error match score distribution computed using grey level image template. Right: MSE match score distribution computed using gradient image template.} 
	\label{fig:mse_dist} 
\end{figure} 

\begin{figure} 
	\centering 
		\includegraphics[width=\textwidth]{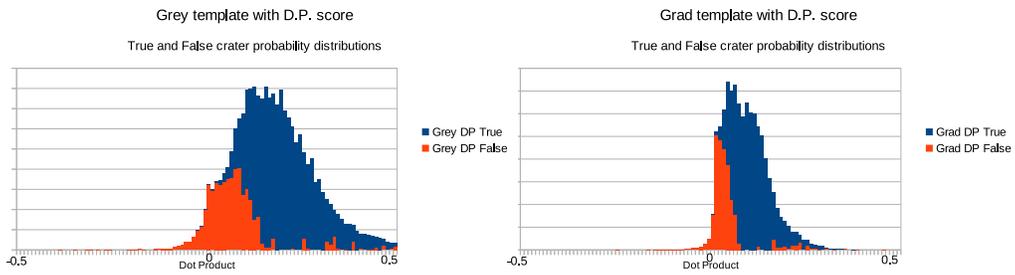} 
	\caption{Left: Dot Product match score distribution computed using grey level image template. Right: DP match score distribution computed using gradient image template.} 
	\label{fig:dp_dist} 
\end{figure}

\begin{figure} 
	\centering 
		\includegraphics[width=0.75\textwidth]{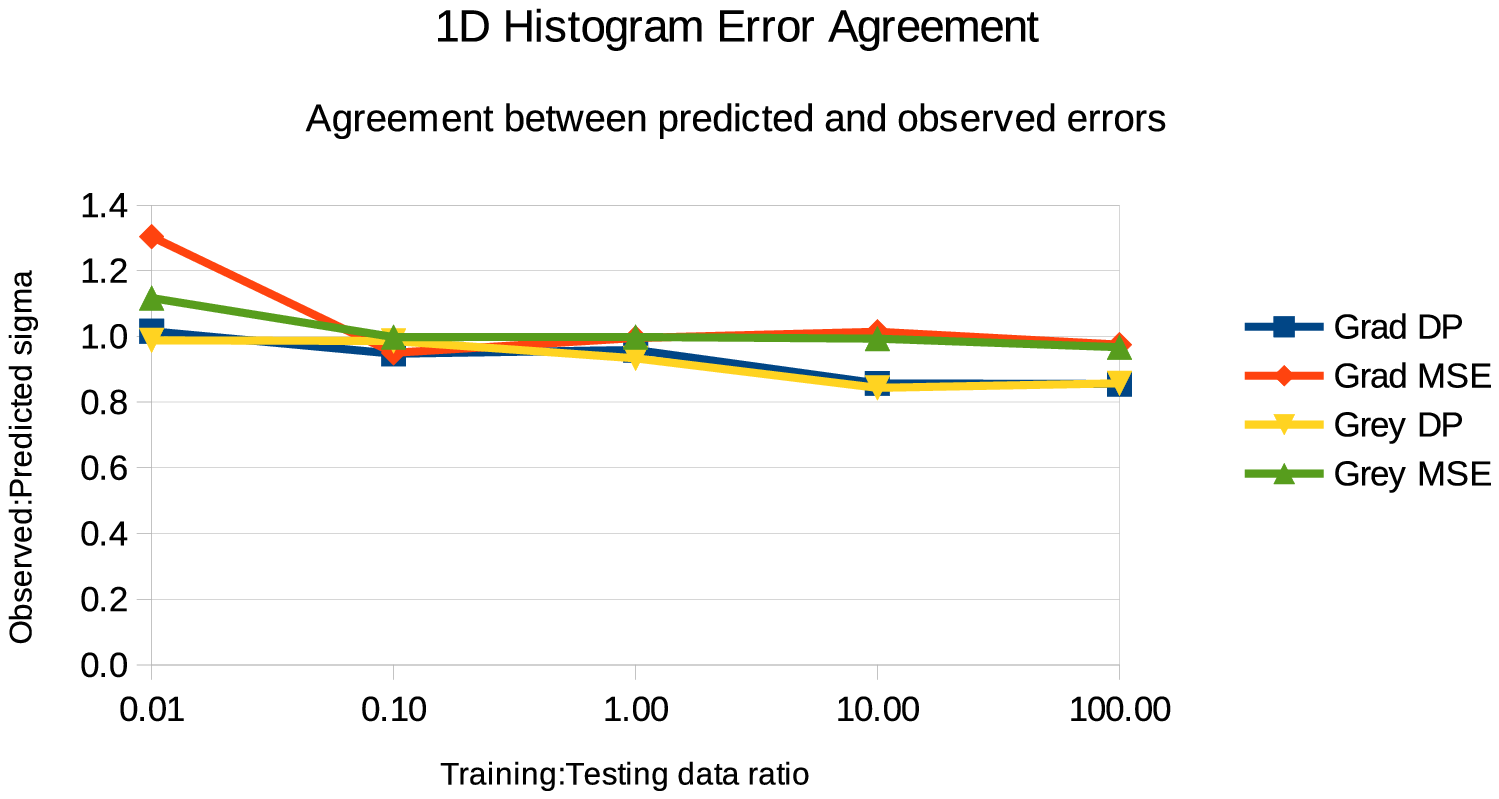} 
	\caption{Corroboration that predicted measurement errors are seen in practice when linear models are constructed and fitted using 1D match score histograms. The x-axis indicates the relative quantities of training and testing data. The y-axis shows observed errors over 1,000 trials per point divided by the predicted errors.} 
	\label{fig:mz_1d_hist_error_agreement} 
\end{figure} 

\begin{figure} 
	\centering 
		\includegraphics[width=0.75\textwidth]{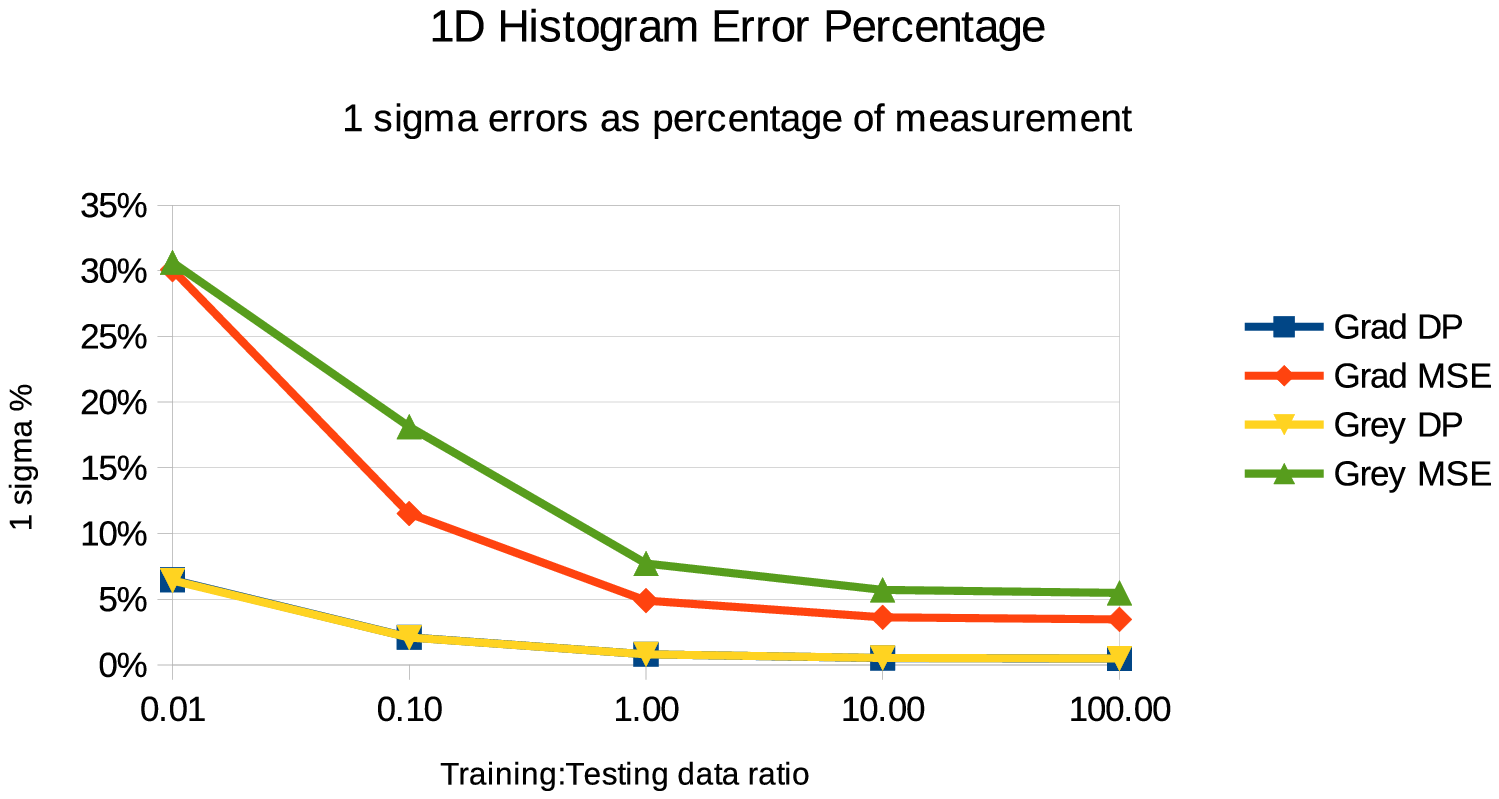} 
	\caption{Measurement errors as percentage of measured quantities when using 1D match score histograms. The x-axis indicates the relative quantities of training and testing data. The y-axis shows one standard deviation of predicted accuracies as a percentage of the measurement.} 
	\label{fig:mz_1d_hist_error_percentage} 
\end{figure} 

\begin{figure} 
	\centering 
		\includegraphics[width=0.75\textwidth]{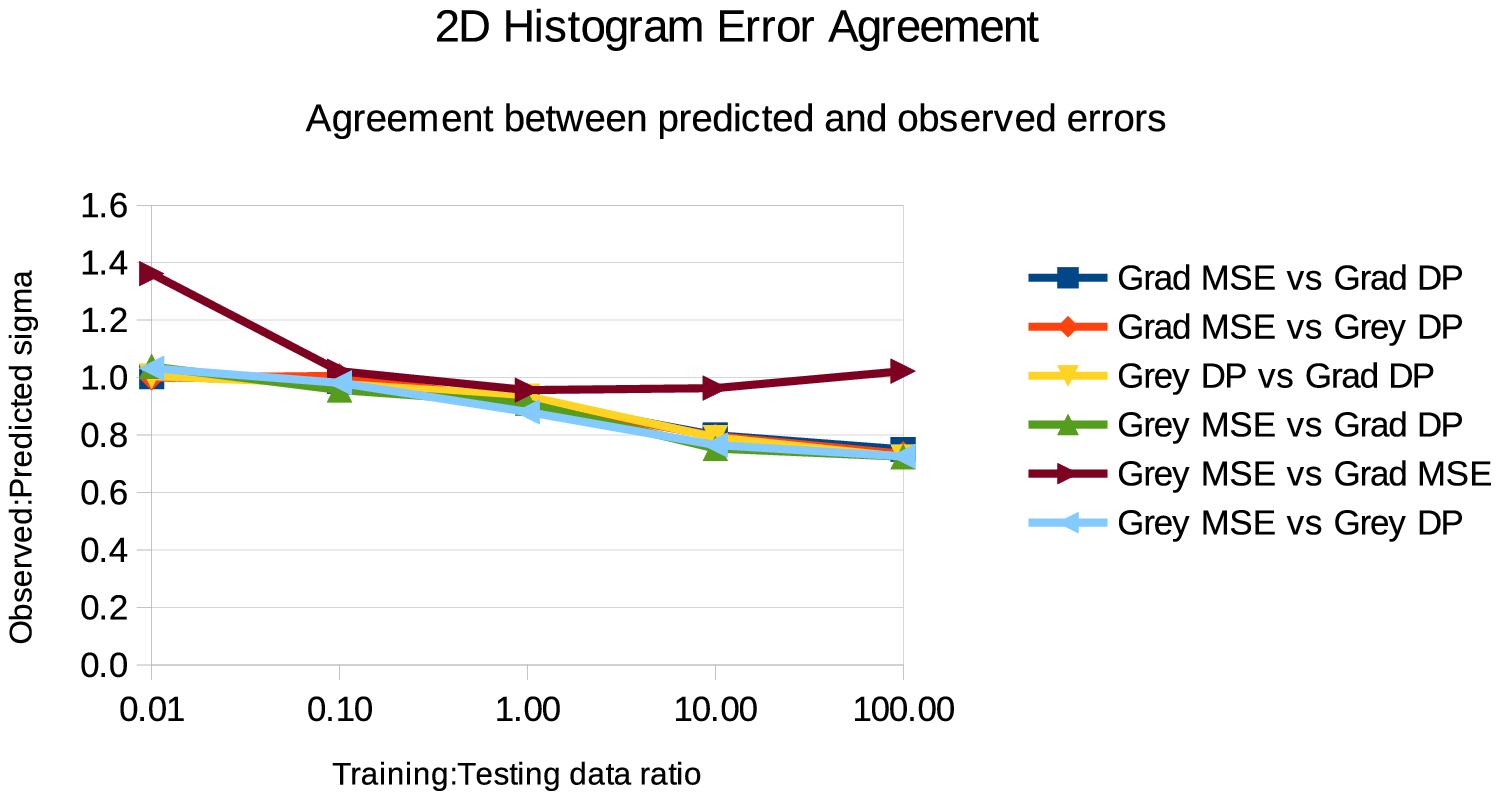} 
	\caption{Corroboration that predicted measurement errors are seen in practice when linear models are constructed and fitted using 2D match score histograms. The x-axis indicates the relative quantities of training and testing data. The y-axis shows observed errors over 1,000 trials per point divided by the predicted errors.} 
	\label{fig:mz_2d_hist_error_agreement} 
\end{figure} 

\begin{figure} 
	\centering 
		\includegraphics[width=0.75\textwidth]{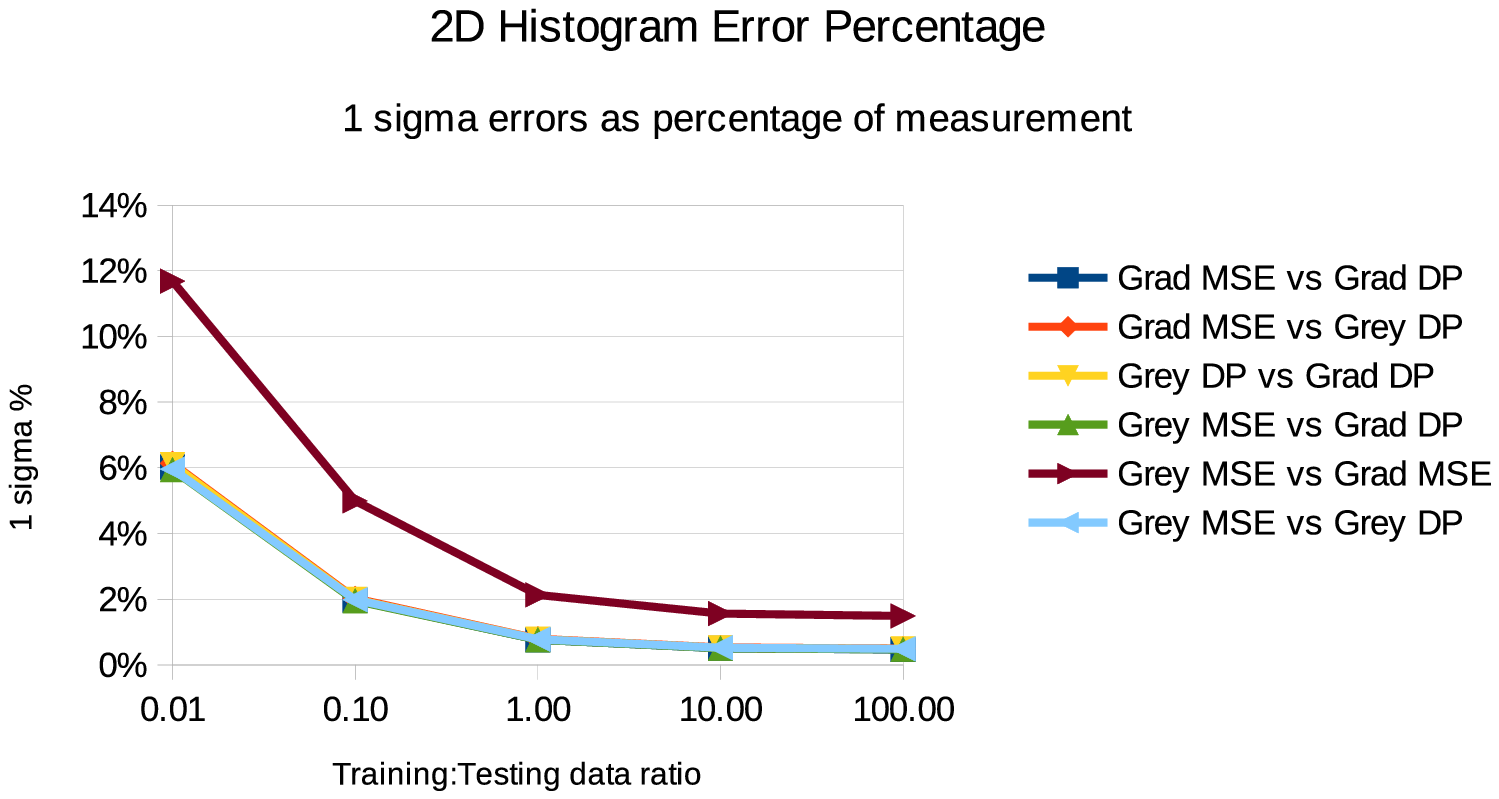} 
	\caption{Measurement errors as percentage of measured quantities when using 2D match score histograms. The x-axis indicates the relative quantities of training and testing data. The y-axis shows one standard deviation of predicted accuracies as a percentage of the measurement.} 
	\label{fig:mz_2d_hist_error_percentage} 
\end{figure}

\begin{figure} 
	\centering 
		\includegraphics[width=0.75\textwidth]{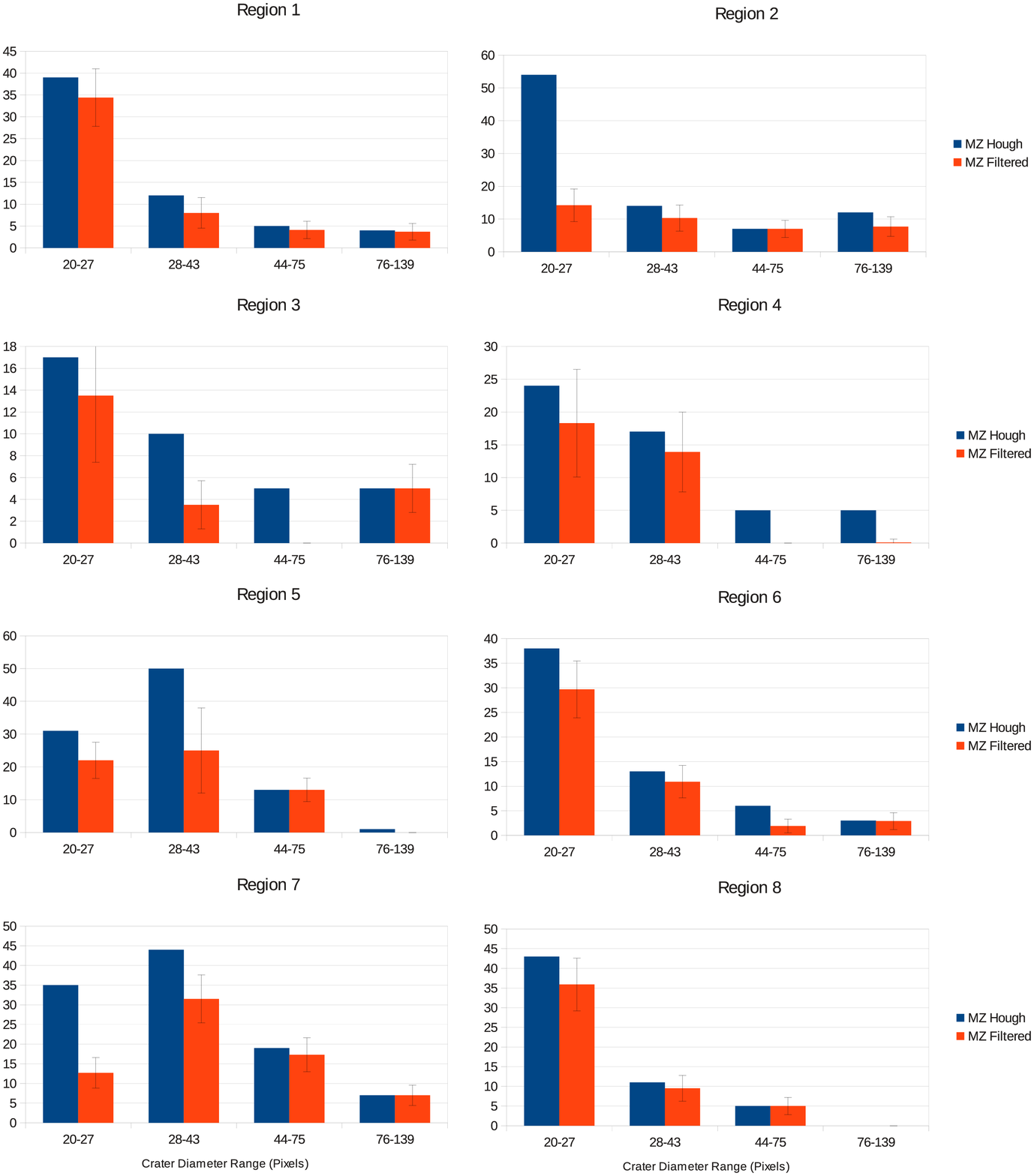} 
	\caption{Blue bars show MoonZoo crater counts including contamination from false positives. Red bars show counts with false positive contamination removed and error bars from the LPM error theory.} 
	\label{fig:hough_vs_filtered} 
\end{figure}

The distribution of match scores (1D cases) for the four combinations of template and similarity measure can be seen in figures \ref{fig:mse_dist} and \ref{fig:dp_dist}. The differently shaded regions correspond to match score values for false positives and ``true'' craters. These are total distribution, taking all candidate craters into consideration. Different spatial subregions exhibit some variation which is accounted for in the LPMs by using 6 PMFs per class (i.e. 6 PMFs for ``true'' craters and an additional 6 for contamination).

Evidence of successfully estimated quantities of ``true'' craters verses contamination can be seen in figures \ref{fig:mz_1d_hist_error_agreement} and \ref{fig:mz_2d_hist_error_agreement}, for 1D and 2D distributions respectively. These show that estimated quantities match ground truth quantities, within predicted errors, i.e. that the predicted accuracies (1 sigma error) match observed accuracies over repeated trails (1,000 per test). The instabilities seen in the 2D plots, where there is deviation away from a flat line at unity, can be explained by underpopulated histogram bins on the left (small data quantities) and growing model discrepancies on the right (large data quantities). However, in all cases the errors were predicted within a half of the actual errors, which for the purposes of plotting error bars successfully addresses the vast majority of the error, thereby still preventing over-interpretation. The actual accuracies attained can be seen in figures \ref{fig:mz_1d_hist_error_percentage} and \ref{fig:mz_2d_hist_error_percentage}, showing that accuracies improve as the quantity of data analysed increases. The best overall performance is achieved when the $S_{DP}$ similarity measure is used.

The reduction in crater counts caused by the removal of contamination within the 8 calibration regions can be seen in figure \ref{fig:hough_vs_filtered}. These plots show the original and corrected counts in narrow size bands, with differential rather than cumulative binning to avoid correlations between error bars. The blue bars show original counts before contamination corrections without any error information and the red bars show contamination removed, with error bars from the LPM error theory.

\subsection{False negative calibration}

\begin{figure} 
	\centering 
		\includegraphics[width=0.75\textwidth]{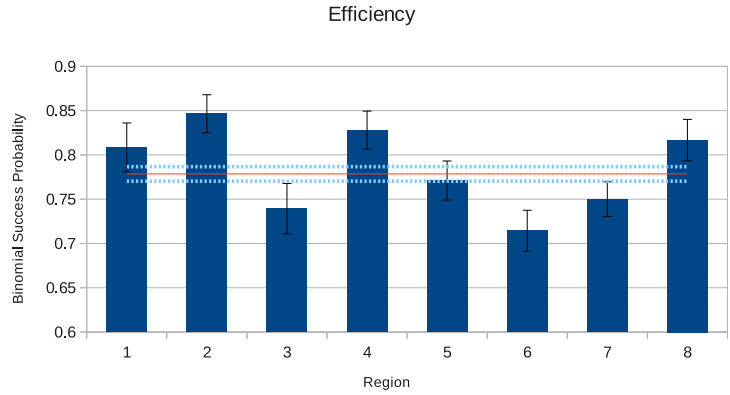} 
	\caption{The Binomial success probabilities of undergraduates identifying craters, i.e. the $P_T$ terms of equ \ref{equ:sfd_count_model}, within the 8 calibration regions. The red horizontal line shows the mean efficiency with dotted blue lines showing 1 standard deviation errors.}
	\label{fig:success_probs} 
\end{figure}

\begin{figure} 
	\centering 
		\includegraphics[width=0.75\textwidth]{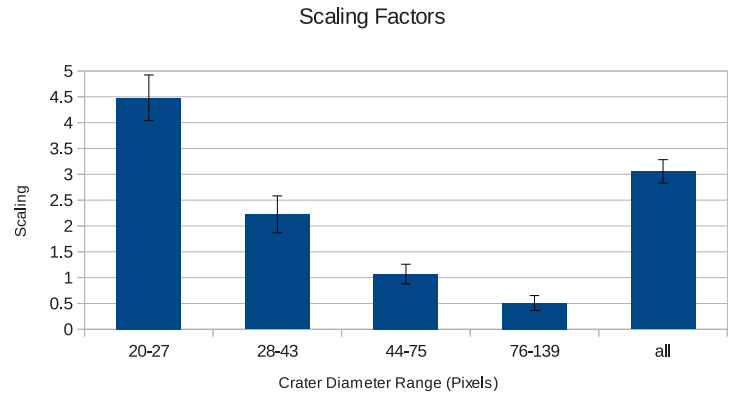} 
	\caption{The scaling factors computed for each calibration region, with the first 4 bars giving size-specific scaling factors and the final bar giving an overall scaling factor.}
	\label{fig:scaling_factors}
\end{figure}

\begin{figure}
	\centering 
		\includegraphics[width=0.75\textwidth]{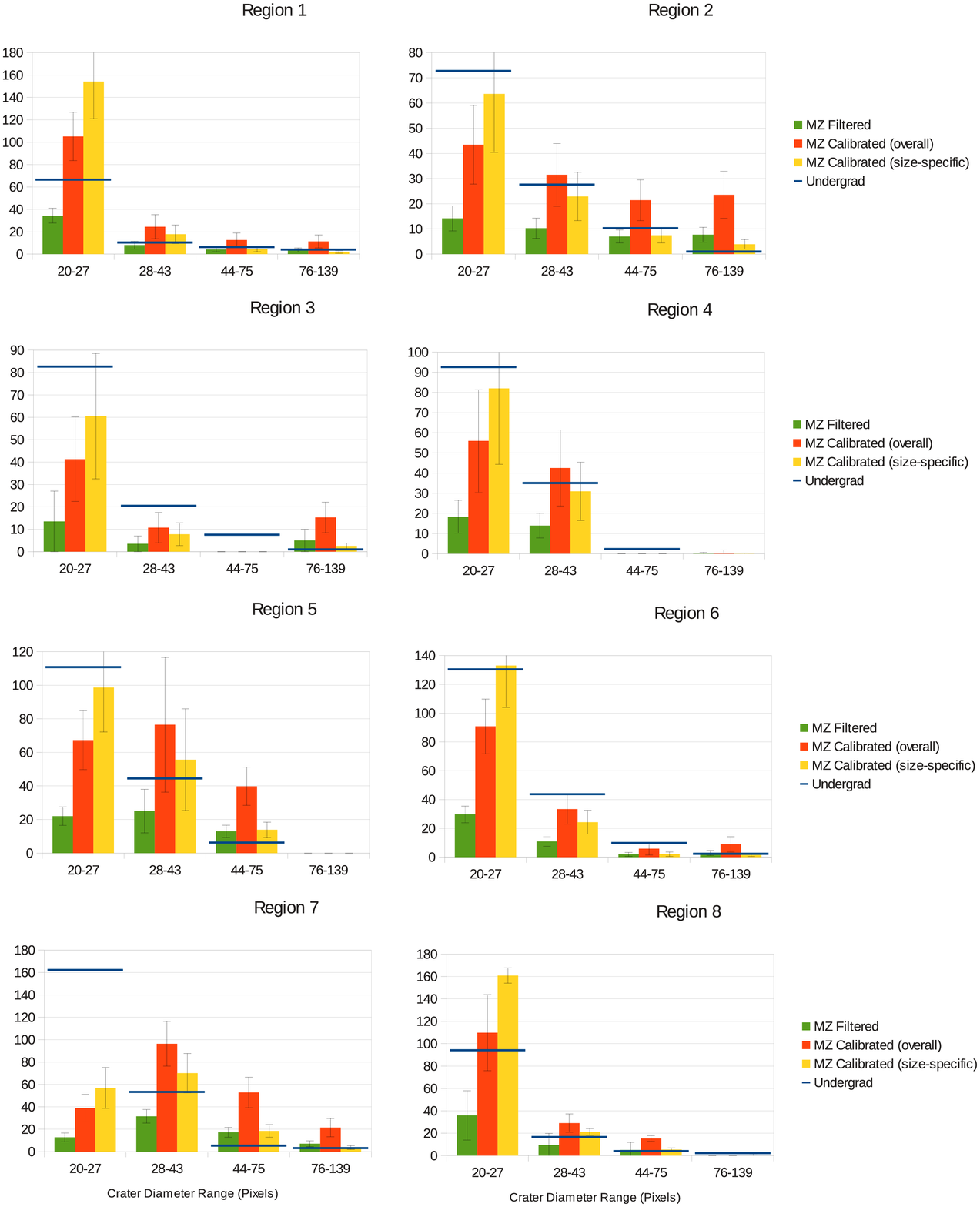} 
	\caption{Calibrated MoonZoo crater counts for the 8 calibration regions. The blue horizontal lines represent the ``ground truth'' undergraduate counts, the green bars are the false positive corrected counts, the red bars are missing data corrected using one over-all scaling factor, and the yellow bars are missing data corrected using size band specific scaling factors. Overall chi-square per degree of freedom values of the three counts to the ground truth are 23.97, 6.20 and 2.88 respectively}
	\label{fig:mz_calibrated} 
\end{figure}

False negative calibration first required a ground-truth definition for comparison. This was gathered from undergraduate counts, with repeatability estimated using Binomial statistics after two counting attempts. Figure \ref{fig:success_probs} shows the estimated Binomial success probabilities of undergraduate crater identification within the 8 calibration regions, i.e. the fraction of craters identified after a single count.

Once defined, the ground truth was used to determine scaling factors required to correct MoonZoo counts in the same regions. Figure \ref{fig:scaling_factors} shows the scaling factors estimated within different size bands and also an overall scaling factor. Figure \ref{fig:mz_calibrated} shows the effect of applying the scaling factors back to the calibration regions as a consistency check. The blue horizontal lines show the undergraduate ground-truth levels with which calibrated counts should be consistent. The uncorrected (green bars), overall corrected (red bars) and size-specific corrected (yellow bars) can be seen to be often significantly away from the ground-truth. Large predicted errors can be seen on most counts.

Chi-square per degree of freedom values (equ. \ref{equ:chi_test}), which should be unity in the case of good conformity, are 23.97, 6.20 and 2.88, respectively for these three cases. The worst discrepancy can be seen at the smaller crater diameters. The value of 2.88 achieved with size-specific scaling is approaching a level at which corrected counts are statistically similar to ground truth, with each count being within only 1 to 2 standard deviations away. 

[Need to also plot standard SFD forms, in line with 1979 recommendations for reference.]

\section{Discussion}

The $N_D = N_T P_T + N_F P_F$ counting model has been proposed to explain the sources of uncertainty observed in real crater counts. The well-established Poisson assumption, which dominates the quantitative analysis of SFDs, has been shown to be invalid for expert, non-expert \cite{Robbins2014} and automated methods making its use unsafe in general cases. This paper has presented a method to address contamination from false positives, thereby correcting for the $N_F P_F$ term, and also calibrating for missing data, thereby correcting for the $N_T (1-P_T)$ false negatives. Evidence has been provided demonstrating the success of the first correction, but successfully correcting for missing data has proved to be challenging for incomplete citizen science datasets. This discussion covers the following points:

\begin{itemize}

\item the successful estimation and correction of false positive contamination, resulting in highly repeatable ``true'' crater counts;

\item the limited success of missing data calibration, resulting in final crater counts with large errors using incomplete MoonZoo data;

\item the implications of these results in terms of conventional Poisson assumptions and comparing independently generated SFDs;

\item and the limitations of the method.

\end{itemize}

\subsection{Successful contamination correction}
\label{sec:success}

MoonZoo annotations contain a fraction of real craters which have been successfully counted, $N_T P_T$, and a fraction of contamination that has been erroneously counted, $N_F P_F$. The contamination correction step has been successfully demonstrated to estimate the amount of these two classes of annotation. The use of LPMs and match score distributions is therefore a feasible method for addressing and reducing the effects of contamination in crater counts. The contamination in MoonZoo data has been successfully measured and corrected for, with contamination levels being regionally dependent, on average affecting around 25\% of the data, as seen in figure \ref{fig:hough_vs_filtered}.

The use of LPMs has been shown to be statistically self-consistent, as quantities were successfully estimated to within predicted levels of accuracy under all tested conditions. Corrected counts were achieved for different types of template, similarity measures and quantities of data. The only significant differences between scenarios is seen in the size of counting errors, with $S_{MSE}$ scores at low quantities of data performing the worst and $S_{DP}$ scores at high quantities performing the best. This can be accounted for through two mechanisms: firstly, the normalised dot product provides greater separability of ``true'' and ``false'' craters, as their distributions overlap less than in the $S_{MSE}$ alternative; secondly, as the quantity of data increases the underlying counting errors naturally improve through the availability of larger samples.

The selection of templates and match scores used are not the only ones available, and no claim is being made that the most successful combination tested is the absolute best possible. However, the best total errors achieved are close to Poisson, as seen in figure \ref{fig:mz_2d_hist_error_percentage}, which is a vast improvement on reported expert repeatability, i.e. up to 7 times worse than Poisson \cite{Robbins2014}. In these best cases the errors on estimated $N_T P_T$ are approximately 1.3 times Poisson: $N_T P_T \pm 1.3 \sqrt{N_T P_T}$, and approach the limit for crater counting accuracies. If these counts did not suffer from missing data, i.e. if $P_T$ was close to 100\%, then the use of expertly trained LPMs could provide a mechanism for producing highly repeatable crater statistics, so long as a single ``expert'' definition could be agreed upon and adopted as a standard.

\subsection{Missing MoonZoo data}

If MoonZoo data was more complete, i.e. $P_T$ was close to 100\%, then the methods presented would provide an effective solution for converting citizen science data into quantitative SFDs. However, the false negative calibration stage revealed large discrepancies between MoonZoo counts and ``expert'' counts, with significantly underestimated counts being found in some regions due to false negatives. The amount of missing data $N_T (1-P_T)$ was too large and too variable for a simple calibration to be performed.

The one-off calibration failed to produce statistically valid counts, as did the diameter dependent calibrations, with respect to both the undergraduate counts and expert counts. The large scaling factors required for corrections resulted in large predicted error bars, which were still underestimates of the true errors leading to poor $\chi^2$ fits. However, an improvement was seen in the $\chi^2$ values as the scaling factors were applied, but more specific calibration would be required for a successful outcome, e.g. degradation specific scaling or a greater number of regional calibration areas. Overall, the MoonZoo data was simply too incomplete and variable for a simple calibration to be achieved. However, this problem could be solved by gathering more data, thereby increasing $P_T$. Calibration regions with known counts can also be introduced ahead of data acquisition, where citizen scientists would be expected meet a minimum level of efficiency before new regions were inspected.

Final errors on processed MoonZoo data (after the application of both correction steps) were upto 50\% or worse, which is less accurate than other reported citizen science repeatability of 20\% to 30\% \cite{Robbins2014}. But, as more data is gathered by MoonZoo less calibration will be required and the near Poisson errors described in section \ref{sec:success} might be achieved.

\subsection{Implications for conventional $\sqrt{N_D}$ error assumption}

In the context of the proposed counting model of equation \ref{equ:sfd_count_model}, the traditional use of Poisson error bars on crater counts assumes that the true crater counting efficiency, $P_T$, is 100\% and that the false crater counting efficiency, $P_F$, is 0\%. In this case all true craters are counted correctly with no contamination, such that the only remaining source of uncertainty is in the natural Poisson cratering fluctuations. If this case can be achieved then $\sqrt{N_D}$ error bars may be applied. However, real crater counts in practice, be they from experts, citizen scientists or algorithms, are not this clean.

If $P_T < 100\%$ and $P_F > 0\%$, and these efficiency terms are themselves variable, then errors on uncorrected counts will be larger than Poisson. This is the general case which can explain the empirical errors reported by others and also the large errors seen in MoonZoo data. Given this is the general case, it is unsafe to compare independently generated SFDs using only Poisson errors. It is also unsafe to trust hypothesis tests and goodness-of-fit tests performed upon SFDs if those tests make the same Poisson assumptions.

However, if $P_T \approx 100\%$ and $P_F > 0\%$, then the false positive correction step (section \ref{sec:false_pos_correction}) can be applied to reduce the effects of contamination resulting in crater counts with near Poisson precision. This case might be approached if citizen science crater data (or results from an algorithm) are highly complete. As long as the dataset contains everything a reasonable expert would identify as being a crater, plus contamination, then that expert can train a LPM to provide Poisson-precision counts from the citizen science datasets consistent with that expert's definition. The results would behave as if that expert performed the complete set of counts personally.

Finally, if efficiencies cannot be controlled or standards achieved then the distribution of $N_D$ must be better understood in terms of the random fluctuations in $P_T$ and $P_F$. The efficiencies, $P_T$ and $P_F$, must be bound between zero (completely missed) and unity (completely counted). These efficiencies must also be random variables, as different users under different viewing conditions and terrains will find it easier or harder to interpret what they see in an image. It might be assumed that these efficiencies peak at some point within their bounds, with peaks which can move and have variable widths. A candidate distribution for describing such random efficiencies is the Beta distribution, which was designed for modelling such uncertain ``success'' probabilities:

\begin{equation}
Beta(p; \alpha, \beta) = \frac{p^{\alpha-1} (1-p)^{\beta-1}}{\int_0^1 u^{\alpha-1} (1-u)^{\beta-1} du}
\end{equation}

where $Beta(p; \alpha, \beta)$ is the probability density of efficiency $p$; and $\alpha$ and $\beta$ are shape parameters. The effects of these terms must be appropriately combined with the Poisson $N$ terms to arrive at a final error model. Much work would be required for this approach to be made practical and it would be difficult to estimate all of the necessary parameters.

\subsection{Limitations}

The simple crater templates used were only designed for small diameters and do not address problems associated with overlapping or nested craters. As such, the method would required better crater models if it were to be used in more complex and densely cratered terrains, such as the Lunar highlands. A range of different templates may also be required if the method is to be extended to a large range of crater sizes to account for changes in morphology.

The use of templates also limits the use of trained LPMs to images with the same solar illumination orientation, as changing shadows change the appearance of templates. It may be time consuming to train alternative LPMs for all required conditions. This should be seen as a semi-automated method, as each new dataset will likely require specific training where a subset of craters are re-examined by an expert in order to provide a representative training subset.

\section{Conclusions}

The key conclusions of this work are as follows:

\begin{enumerate}

\item Previous work by others \cite{Robbins2014} has shown that the conventional Poisson error assumption on crater counts does not hold true, as repeatability studies show true errors up to 7 times worse than Poisson. It is therefore logically defensible to suggest an alternative counting model, $N_D = N_T P_T + N_F P_F$, which explicitly addresses other sources of uncertainty, including false positives and false negatives.

\item The effects of false positive contamination, $N_F P_F$, can be successfully quantified and corrected for using Linear Poisson Models, utilising objective differences in template crater match scores, so long as an appropriate standard template and ground truth can be established. Furthermore, statistically consistent results can be achieved for different types of templates and match scores.

\item MoonZoo crater counts contain approximately 25\% contamination, varying from region to region.

\item It is more difficult to account for missing data via simple calibration. MoonZoo data is highly incomplete in many regions, but gathering more data and controlling efficiencies could be a simple solution to the problem. Data should be gathered in such a way that $P_T$ approaches 100\% and that this efficiency is approximately constant for all regions.

\item Whilst it is unsafe in general to compare SFDs using $\sqrt{N_D}$ errors, steps might be taken, such as the adoption of standards and use of LPMs for contamination corrections, allowing errors to approach this level, but these steps are not trivial. If applied, evidence presented in this paper suggests crater counts could be achieved to within 1.3 times Poisson errors - a large improvement upon uncorrected counts.

\item If standards are not adopted then it will remain difficult for independently produced SFDs to be compared within good levels of accuracy. Meaningful comparisons between SFDs may be restricted to those produced under equivalent conditions, e.g. by the same expert using the same interface, minimising the variability of the $P_T$ and $P_F$ efficiency terms. A more comprehensive error theory may also be required to incorporate these efficiency variations for more general comparisons to be made.

\item The proposed methods are currently limited to small craters in sparsely cratered regions, but work could be undertaken to widen applicability.

\end{enumerate}

Overall, this paper provides new insights into the sources of crater counting uncertainties, and in particular the large uncertainties present in citizen science data. Some key recommendations of \cite{CraterWorkingGroup} have been revisited in order to emphasis the importance of error analysis, goodness-of-fits and the avoidance of subjectivity.

\section*{Acknowledgements}
We thank Sean Corrigan, Alex Griffiths, Tim Gregory, Hazel Blake, Dayl Martin, Maggie Sliz, Joe Scaife and Pavel Kamenov for their assistance in providing ground-truth crater counts. We'd also like to thank the Leverhulme Trust for providing project funding (RPG-2014-019).

\bibliographystyle{plain}
\bibliography{refs}

\end{document}